\documentstyle[11pt]{article}

\newenvironment{dispquo}%
{\vspace{\topsep}\begin{quote}}%
{\end{quote}\vspace{\topsep}}

\newenvironment{dispeqn}%
{\vspace{\topsep}\begin{eqnarray*}}%
{\end{eqnarray*}\vspace{\topsep}}

\newenvironment{reverseindent}%
{\begin{list}{}{\setlength{\labelsep}{0in}
	        \setlength{\labelwidth}{0in}
	        \setlength{\itemindent}{-\leftmargin}}}%
{\end{list}}

\begin{document}

\begin{center}
{\Large\bf Qualitative and Quantitative Models\\
of Speech Translation}\\[5mm]
{\large Hiyan Alshawi}\\[2mm]
AT\&T Bell Laboratories\\
600 Mountain Ave.,
Murray Hill, New Jersey 07974\\[1mm]
{\tt hiyan@research.att.com}\\[2mm]
Archive number: cmp-lg/9408014\\[3mm]
{\bf ABSTRACT}
\end{center}
\noindent
This paper compares a qualitative reasoning model of
translation with a quantitative statistical model.
We consider these models within the context of two hypothetical
speech translation systems, starting with a logic-based design
and pointing out which of its characteristics are best
preserved or eliminated in moving to the second,
quantitative design. The quantitative language and
translation models are based on relations between lexical heads of
phrases. Statistical parameters for structural dependency, lexical
transfer, and linear order are used to select a
set of implicit relations between words in a source utterance,
a corresponding set of relations between target language words,
and the most likely translation of the original utterance.

\section{Introduction}

In recent years there has been a resurgence of interest in
statistical approaches to natural language processing.
Such approaches are not new, witness the statistical approach
to machine translation suggested by Weaver (1955), but the
current level of interest is largely due to the success of applying
hidden Markov models and N-gram language models in speech
recognition.
This success was directly measurable in terms of word recognition
error rates, prompting language processing
researchers to seek corresponding
improvements in performance and robustness. A speech translation
system, which by necessity combines speech and language technology,
is a natural place to consider combining the
statistical and conventional approaches and much of this paper
describes probabilistic models of structural
language analysis and translation.
Our aim will be to provide an overall model for translation
with the best of both worlds. Various factors will lead us to
conclude that a lexicalist statistical model with dependency
relations is well suited to this goal.

As well as this quantitative approach, we will consider
a constraint/logic based approach and try to distinguish
characteristics that we wish to preserve from those that
are best replaced by statistical models.
Although perhaps implicit in many conventional approaches to
translation, a characterization in logical terms of what
is being done is rarely given, so we will attempt to make
that explicit here, more or less from first principles.

Before proceeding, I will first examine some fashionable
distinctions in section~\ref{distinctions} in order to
clarify the issues involved in comparing these approaches.
I will attempt to argue that the important
distinction is not so much a rational-empirical
or symbolic-statistical distinction but rather a
qualitative-quantitative one.
This is followed by discussion of the logic-based model
in section~\ref{logical}, the overall quantitative model
in section~\ref{arch}, monolingual models in section~\ref{lm},
translation models in section~\ref{tm}, and some
conclusions in section~\ref{compare}.
We concentrate throughout on what information about language
and translation is coded and how it is expressed as logical
constraints or statistical parameters. Although important, we
will say little about search algorithms, rule acquisition,
or parameter estimation.

\section{Qualitative and Quantitative Models}
\label{distinctions}

One contrast often taken for granted is the identification of
a `statistical-symbolic' distinction in language processing
as an instance
of the empirical vs. rational debate. I believe this contrast
has been exaggerated though historically it has had some validity
in terms of accepted practice. Rule based
approaches have become more empirical in a number of ways:
First, a more empirical approach is being adopted to grammar
development whereby the rule set is modified according to
its performance against corpora of natural text
(e.g. Taylor, Grover, and Briscoe 1989).
Second, there is a class of
techniques for learning rules from text, a recent example
being Brill 1993. Conversely, it is possible
to imagine building a language model in which all
probabilities are estimated according to intuition
without reference to any
real data, giving a probabilistic model that is not empirical.

Most language processing labeled as statistical involves associating
real-number valued parameters to configurations of symbols.
This is not surprising given that natural language, at least
in written form, is explicitly symbolic. Presumably, classifying
a system as symbolic must refer to a different set of (internal)
symbols, but even this does not rule out many statistical systems
modeling events involving nonterminal categories and
word senses. Given that the notion of a symbol, let alone
an `internal symbol', is itself a slippery one, it
may be unwise to build our theories of language, or even the way we
classify different theories, on this notion.

Instead, it would seem that the real contrast
driving the shift towards statistics in
language processing is a contrast between {\it qualitative} systems
dealing exclusively with combinatoric constraints,
and {\it quantitative} systems that involve computing
numerical functions.
This bears directly on the problems of brittleness and complexity that
discrete approaches to language processing share with,
for example, reasoning
systems based on traditional logical inference. It relates
to the inadequacy of the dominant theories in linguistics
to capture `shades' of meaning or degrees of acceptability
which are often recognized by people outside the field as
important inherent properties of natural language.
The qualitative-quantitative distinction can also be seen as
underlying the difference between classification systems based
on feature specifications, as used in unification formalisms
(Shieber 1986),
and clustering based on a variable degree of
granularity (e.g. Pereira, Tishby and Lee 1993).

It seems unlikely that these continuously variable aspects
of fluent natural language can be captured by a purely
combinatoric model. This naturally leads to the question
of how best to introduce quantitative modeling into
language processing.
It is not, of course, necessary for the quantities of a
quantitative model to be probabilities. For
example, we may wish to define real-valued functions
on parse trees that reflect the extent to which
the trees conform to, say, minimal attachment and parallelism
between conjuncts. Such functions have been used in tandem
with statistical functions in experiments on
disambiguation (for instance Alshawi and Carter 1994).
Another example is connection strengths in neural network
approaches to language processing, though it has
been shown that certain networks are effectively computing
probabilities (Richard and Lippmann 1991).

Nevertheless, probability theory does offer a coherent
and relatively well understood framework for selecting
between uncertain alternatives, making it a natural choice
for quantitative language processing. The case for
probability theory is strengthened by a well
developed empirical methodology in the form of statistical
parameter estimation. There is
also the strong connection between probability theory
and the formal theory of information and communication,
a connection that has been
exploited in speech recognition, for example using the
concept of entropy to provide a motivated way of
measuring the complexity of a recognition problem
(Jelinek et al. 1992).

Even if probability theory remains, as it currently is,
the method of choice in making language processing
quantitative, this still leaves the field wide open
in terms of carving up language processing into
an appropriate set of events for probability theory to
work with. For translation, a very direct approach
using parameters based on surface positions of words
in source and target sentences was adopted
in the Candide system (Brown et al. 1990). However, this
does not capture important structural properties of
natural language. Nor does
it take into account generalizations about translation
that are independent of the exact word order in source
and target sentences. Such generalizations are, of
course, central to qualitative structural approaches to
translation (e.g. Isabelle and Macklovitch 1986, Alshawi
et al. 1992).

The aim of the quantitative language and translation models
presented in sections~\ref{lm} and \ref{tm} is to employ
probabilistic parameters that reflect linguistic structure
without discarding rich lexical information or making
the models too complex to train automatically. In terms
of a traditional classification, this would be seen
as a `hybrid symbolic-statistical' system because it deals with
linguistic structure. From our perspective,
it can be seen as a quantitative version of the logic-based
model because both models attempt to capture similar
information (about the organization of words into phrases
and relations holding between these phrases or their referents),
though the tools of modeling are substantially different.

\section{Dissecting a Logic-Based System}
\label{logical}

We now consider a hypothetical speech translation
system in which the language processing components follow
a conventional qualitative transfer design. Although hypothetical,
this design and its components are similar to those
used in existing database query (Rayner and Alshawi 1992)
and translation systems (Alshawi et al 1992).
More recent versions of these systems have been gradually taking on
a more quantitative flavor, particularly with respect to
choosing between alternative analyses, but our hypothetical
system will be more purist in its qualitative approach.

The overall design is as follows.
We assume that a speech recognition subsystem delivers a
list of text strings corresponding to transcriptions of an
input utterance. These recognition hypotheses are passed
to a parser which applies a logic-based grammar and lexicon
to produce a set of logical forms,
specifically formulas in first order logic corresponding to
possible interpretations of the utterance. The logical forms
are filtered by contextual and word-sense constraints,
and one of them is passed to the translation component.
The translation relation is expressed by a set of first
order axioms which are used by a theorem prover to derive
a target language logical form that is equivalent (in some
context) to the source logical form. A grammar for the
target language is then applied to the target form, generating
a syntax tree whose fringe is passed to a speech
synthesizer.

Taking the various components in turn, we make a note
of undesirable properties that might be improved
by quantitative modeling.

\subsection*{Analysis and Generation}

A grammar, expressed as a set of syntactic rules (axioms) $G_{syn}$ and
a set of semantic rules (axioms) $G_{sem}$ is used to support
a relation ${form}$ holding between strings $s$ and
logical forms $\phi$ expressed in first order logic:
\begin{dispquo}
$ G_{syn} \cup G_{sem} \models {form}(s,\phi)$.
\end{dispquo}
The relation ${form}$ is many-to-many, associating a string with
linguistically possible logical form interpretations. In the analysis
direction, we are given $s$ and search for logical forms $\phi$,
while in generation we search for strings $s$ given $\phi$.

For analysis and generation, we are treating strings $s$
and logical forms $\phi$ as object level entities. In
interpretation and translation, we will move down from this meta-level
reasoning to reasoning with the logical forms as propositions.

The list of text strings handed by the recognizer to the
parser can be assumed to be ordered in accordance with some
acoustic scoring scheme internal to the recognizer.
The magnitude of the scores is ignored by
our qualitative language processor; it simply processes
the hypotheses one at a time until it finds one for which
it can produce a complete logical form interpretation
that passes grammatical and interpretation constraints,
at which point it discards the remaining hypotheses.
Clearly, discarding the acoustic score and taking the
first hypothesis that satisfies the constraints
may lead to an interpretation that
is less plausible than one derivable from a hypothesis
further down in the recognition list. But there is no
point in processing these later hypotheses since we will
be forced to select one interpretation essentially at random.

\paragraph{Syntax}

The syntactic rules in $G_{syn}$ relate `category'
predicates $c_0, c_1, c_2$
holding of a string and two spanning substrings (we limit the
rules here to two daughters for simplicity):
\begin{dispquo}
$ c_0(s_0) \wedge daughters(s_0, s_1, s_2) \leftarrow \\
  \hspace*{5mm}
  c_1(s_1) \wedge c_2(s_2) \wedge (s_0=concat(s_1,s_2))$
\end{dispquo}
(Here, and subsequently, variables like $s_0$ and $s_1$ are
implicitly universally quantified.)
$G_{syn}$ also includes lexical axioms for particular strings $w$
consisting of single words:
\begin{dispquo}
$ c_1(w) $, \hspace*{8mm} \dots  \hspace*{8mm} $ c_m(w) $.
\end{dispquo}
For a feature-based grammar, these rules can include
conjuncts constraining the values, $a_1, a_2, \dots$,
of discrete-valued functions $f$
on the strings:
\begin{dispquo}
$ f(w) = a_1, \hspace*{8mm} f(s_0)=f(s_1)$.
\end{dispquo}

The main problem here is that such grammars have
no notion of a degree of grammatical acceptability -- a sentence
is either grammatical or ungrammatical. For small grammars
this means that perfectly acceptable strings are often
rejected; for large grammars we get a vast number of
alternative trees so the chance of selecting the correct tree
for simple sentences can get worse as the grammar coverage
increases. There is also the problem of requiring increasingly
complex feature sets to describe idiosyncrasies in the lexicon.

\paragraph{Semantics}

Semantic grammar axioms belonging to $G_{sem}$ specify a `composition'
function $g$ for deriving a logical form for a phrase from those
for its subphrases:
\begin{dispquo}
$ {form}(s_0, g(\phi_1,\phi_2)) \leftarrow \\
  \hspace*{10mm}
  daughters(s_0, s_1, s_2) \wedge c_1(s_1) \wedge c_2(s_2) \wedge c_0(s_0) \\
  \hspace*{20mm}
  \wedge {form}(s_1,\phi_1) \wedge {form}(s_2,\phi_2)$
\end{dispquo}
The interpretation rules for strings
bottom out in a set of lexical semantic rules associating
words with predicates ($p_1, p_2, \dots$) corresponding to
`word senses'. For a particular word and syntactic category, there
will be a (small, possibly empty) finite set of such word sense
predicates:
\begin{dispquo}
$c_i(w) \rightarrow {form}(w,p^i_1)$\\
\dots\\
$c_i(w) \rightarrow  {form}(w,p^i_m)$.
\end{dispquo}

First order logic was assumed as the semantic representation
language because it comes with well understood, if not very
practical, inferential machinery for constraint solving.
However, applying this machinery requires
making logical forms fine grained to a degree often
not warranted by the information the speaker of an
utterance intended to convey.
An example of this is explicit scoping which leads (again) to
large numbers of alternatives which the qualitative model
has difficulty choosing between. Also,
many natural language sentences cannot be expressed in
first order logic without resort to elaborate formulas
requiring complex semantic composition rules.
These rules can be simplified by
using a higher order logic but
at the expense of even less practical inferential machinery.

In applying the grammar in generation we are faced with the problem
of balancing over and under-generation by tweaking
grammatical constraints, there being no way to prefer
fully grammatical target sentences over more marginal ones.
Qualitative approaches to grammar tend to emphasize the ability
to capture generalizations as the main measure of success in
linguistic modeling. This might explain why producing appropriate
lexical collocations is rarely addressed seriously in these models,
even though lexical collocations are important for
fluent generation.
The study of collocations for generation fits in more
naturally with statistical techniques, as illustrated
by Smajda and McKeown (1990).

\subsection*{Interpretation}

In the logic-based model, interpretation is the process of
identifying from the possible interpretations $\phi$ of
$s$ for which ${form}(s,\phi)$ hold, ones that are consistent with
the context of interpretation. We can state this as follows:
\begin{dispquo}
$  R \cup S \cup A \models \phi$.
\end{dispquo}
Here, we have separated the context into a contingent set of
contextual propositions $S$ and a set $R$ of
(monolingual) `meaning postulates', or
selectional restrictions, that constrain the word sense predicates
in all contexts. $A$ is a set of assumptions sufficient to
support the interpretation $\phi$ given $S$ and $R$. In other words,
this is `interpretation as abduction' (Hobbs et al. 1988), since
abduction, not deduction, is needed to arrive at the assumptions $A$.

The most common types of
meaning postulates in $R$ are those for restriction, hyponymy,
and disjointness, expressed as follows:
\begin{dispquo}
$p_1(x_1,x_2) \rightarrow p_2(x_1)$ \hspace*{4mm}restriction;\\
$p_2(x) \rightarrow p_3(x)$ \hspace*{4mm}hyponymy;\\
$\neg (p_3(x) \wedge  p_4(x))$ \hspace*{4mm}disjointness.
\end{dispquo}
Although there are compilation techniques (e.g. Mellish 1988)
which allow selectional constraints stated in this fashion to
be implemented efficiently, the scheme is problematic
in other respects. To start with, the assumption of
a small set of senses for a word is at best
awkward because it is difficult to arrive at an optimal
granularity for sense distinctions. Disambiguation with
selectional restrictions expressed as meaning postulates
is also problematic because it is virtually impossible
to devise a set of postulates that will always filter
all but one alternative. We are thus forced to
under-filter and make an arbitrary choice between
remaining alternatives.

\subsection*{Logic based translation}

In both the quantitative and qualitative models we
take a transfer approach to translation. We do not
depend on interlingual symbols, but instead map a
representation with constants associated with
the source language into a corresponding expression
with constants from the target language. For the qualitative
model, the operable notion of correspondence is based on
logical equivalence and the constants are source word
sense predicates $p_1, p_2, \dots$ and
target sense predicates $q_1, q_2, \dots$.

More specifically, we
will say the translation relation between a source logical
form $\phi_s$ and a target logical form $\phi_t$ holds
if we have
\begin{dispquo}
$ B \cup S \cup A' \models (\phi_s \leftrightarrow \phi_t) $
\end{dispquo}
where $B$ is a set of monolingual and bilingual meaning postulates,
and $S$ is a set of formulas characterizing
the current context.
$A'$ is a set of assumptions that includes the
assumptions $A$ which supported $\phi_s$. Here bilingual
meaning postulates are first order axioms relating source
and target sense predicates. A typical bilingual postulate
for translating between $p_1$ and $q_1$ might be of the form:
\begin{dispquo}
$p_5(x_1) \rightarrow (p_1(x_1,x_2) \leftrightarrow q_1(x_1,x_2))$.
\end{dispquo}

The need for the assumptions
$A'$ arises when a source language word
is vaguer that its possible translations in the target
language, so different choices of target words will
correspond to translations under different assumptions.
For example, the condition $p_5(x_1)$ above might be proved
from the input logical form, or it might need to be assumed.

In the general case, finding solutions (i.e. $A',\phi_t$ pairs)
for the abductive schema is an undecidable theorem proving
problem. This can be alleviated by placing restrictions
on the form of meaning postulates and input formulas and
using heuristic search methods. Although such an approach
was applied with some success in a limited-domain
system translating logical forms
into database queries (Rayner and Alshawi 1992), it is
likely to be impractical for language translation with
tens of thousands of sense predicates and related axioms.

Setting aside the intractability issue, this approach
does not offer a principled way of choosing between
alternative solutions proposed by the prover.
One would like to prefer solutions with `minimal' sets of
assumptions, but it is difficult to find motivated
definitions for this minimization in a purely qualitative
framework.

\section{Quantitative Model Components}
\label{arch}

\subsection{Moving to a Quantitative Model}

In moving to a quantitative architecture, we propose to
retain many of the basic characteristics of the qualitative model:
\begin{itemize}

\item A transfer organization with analysis, transfer, and
      generation components.

\item Monolingual models that can be used for both
      analysis and generation.

\item Translation models that exclusively code contrastive
      (cross-linguistic) information.

\item Hierarchical phrases capturing recursive linguistic
      structure.

\end{itemize}

Instead of feature based syntax trees and first-order
logical forms we will adopt a simpler, monostratal representation
that is more closely related to those found in
dependency grammars (e.g. Hudson 1984). Dependency representations
have been used in large scale qualitative machine translation
systems, notably by McCord (1988). The notion of a lexical `head'
of a phrase is central to these representations because
they concentrate on relations between such lexical heads.
In our case, the dependency representation is monostratal in that
the relations may include ones normally classified as belonging
to syntax, semantics or pragmatics.

One salient property of our language model is that it is strongly
lexical:
it consists of statistical parameters associated with relations
between lexical items and the number and ordering of dependents
of lexical heads. This lexical anchoring facilitates statistical
training and sensitivity to lexical variation and collocations.
In order to gain the benefits of probabilistic modeling, we
replace the task of
developing large rule sets with the task of estimating
large numbers of statistical parameters for the monolingual and
translation models. This gives rise to a new
cost trade-off in human annotation/judgement versus barely
tractable fully automatic training. It also necessitates further
research on lexical similarity and clustering (e.g. Pereira, Tishby
and Lee 1993, Dagan, Marcus and Markovitch 1993) to improve parameter
estimation from sparse data.

\subsection*{Translation via Lexical Relation Graphs}

The model associates phrases with {\it relation graphs}.
A relation graph is a directed labeled graph
consisting of a set of {\it relation edges}.
Each edge has the form of an atomic proposition
\begin{dispquo}
$r(w_i,w_j)$
\end{dispquo}
where $r$ is a relation symbol, $w_i$ is the lexical head
of a phrase and $w_j$ is the lexical head of another
phrase (typically a subphrase of the phrase headed by $w_i$).
The nodes $w_i$ and $w_j$ are word occurrences
representable by a word and an index,
the indices uniquely identifying particular
occurrences of the words in a discourse or corpus.
The set of relation symbols is open ended, but the first argument
of the relation is always interpreted as the {\it head} and
the second as the {\it dependent} with respect to this relation.
The relations in the models for the source and target languages
need not be the same, or even overlap.
To keep the language models simple,
we will mainly restrict ourselves here to dependency
graphs that are trees with unordered siblings. In particular,
phrases will always be contiguous strings of words and dependents
will always be heads of subphrases.

Ignoring algorithmic issues relating to compactly representing
and efficiently searching the space of alternative hypotheses,
the overall design of the quantitative system is as follows.
The speech recognizer produces a set of word-position hypotheses
(perhaps in the form of a word lattice) corresponding to a
set of string hypotheses for the input. The
source language model is used to compute a set of possible
relation graphs, with associated probabilities, for each string
hypothesis.
A probabilistic graph translation model then provides,
for each source relation graph, the probabilities of deriving
corresponding graphs with word occurrences from the target
language. These target graphs include all the words of possible
translations of the utterance hypotheses but do not specify
the surface order of these words. Probabilities for different
possible word orderings are computed according to ordering
parameters which form part of the target language model.

In the following section we explain
how the probabilities for these various processing stages are combined
to select the most likely target word sequence. This word sequence
can then be handed to the speech synthesizer. For
tighter integration between generation and synthesis,
information about the derivation of the target utterance can also
be passed to the synthesizer.

\subsection{Integrated Statistical Model}
\label{overstat}

The probabilities associated with phrases in the above description
are computed according to the
statistical models for analysis, translation, and generation.
In this section we show the relationship between these
models to arrive at an overall statistical
model of speech translation.
We are not considering training issues in this paper,
though a number of now familiar techniques ranging from
methods for maximum likelihood estimation to direct
estimation using fully annotated data are applicable.

The objects involved in the overall model are as follows
(we omit target speech synthesis under the assumption
that it proceeds deterministically from a target
language word string):
\begin{itemize}
\item $A_s$: (acoustic evidence for) source language speech
\item $W_s$: source language word string
\item $W_t$: target language word string
\item $C_s$: source language relation graph
\item $C_t$: target language relation graph
\end{itemize}

Given a spoken input in the source language, we wish to
find a target language string that is the most likely
translation of the input. We are thus interested in
the conditional probability of $W_t$ given $A_s$.
This conditional probability can be expressed as follows
(cf. Chang and Su 1993):
\begin{dispeqn}
\lefteqn{ P(W_t | A_s) = } \\
&\sum_{W_s,C_s,C_t} & P(W_s | A_s) \: P(C_s | W_s,A_s) \\
&                   & P(C_t | C_s,W_s,A_s) \: P(W_t | C_t,C_s,W_s,A_s).
\end{dispeqn}

We now apply some simplifying independence assumptions concerning
relation graphs. Specifically, that their derivation from
word strings is independent of acoustic information; that their
translation is independent of the original words and acoustics
involved; and that target word string generation from
target relation edges is independent of the source language
representations. The extent to which these (Markovian) assumptions
hold depend on the extent to which relation edges represent all
the relevant information for translation. In particular it
means they should express aspects of surface relevant to meaning,
such as topicalization, as well as predicate argument structure.
In any case, the simplifying assumptions give the following:
\begin{dispeqn}
\lefteqn {P(W_t | A_s) \simeq} \\
& \sum_{W_s,C_s,C_t} P(W_s | A_s) \: P(C_s | W_s)
                       \: P(C_t | C_s) \: P(W_t | C_t).
\end{dispeqn}
This can be rewritten with two applications of Bayes rule:
\begin{dispeqn}
&\sum_{W_s,C_s,C_t} & P(A_s | W_s) \: (1/P(A_s)) \: P(W_s | C_s) \\
&                   & P(C_s) \: P(C_t | C_s) \: P(W_t | C_t).
\end{dispeqn}

Since $A_s$ is given, $1/P(A_s)$ is a constant which can be
ignored in finding the maximum of $P(W_t | A_s)$. Determining
$W_t$ that maximizes $P(W_t | A_s)$ therefore involves the
following factors:

\begin{itemize}
\item $P(A_s | W_s)$: source language acoustics
\item $P(W_s | C_s)$: source language generation
\item $P(C_s)$:       source content relations
\item $P(C_t | C_s)$: source to target transfer
\item $P(W_t | C_t)$: target language generation
\end{itemize}

We assume that the speech recognizer provides acoustic
scores proportional to $P(A_s | W_s)$ (or logs thereof). Such
scores are normally computed by speech recognition systems,
although they are usually also multiplied
by word-based language model probabilities $P(W_s)$ which
we do not require in this application context.
Our approach to language modeling, which covers the
content analysis and language generation factors, is presented
in section~\ref{lm} and the transfer probabilities fall
under the translation model of section~\ref{tm}.

Finally note that by another application of Bayes rule we
can replace the two factors $P(C_s) P(C_t | C_s)$ by
$P(C_t) P(C_s | C_t)$ without changing other parts of the
model.  This latter formulation allows us to apply
constraints imposed by the target language model to
filter inappropriate possibilities suggested by analysis
and transfer. In some respects this is similar to
Dagan and Itai's (1994) approach to word sense disambiguation
using statistical associations in a second language.

\section{Language Models}
\label{lm}

\subsection{Language Production Model}

Our language model can be viewed in terms of a probabilistic
generative process based on the choice of lexical `heads' of
phrases and the recursive generation of subphrases and their
ordering. For this purpose, we can define the head word of a
phrase to be the word that most strongly influences the way the
phrase may be combined with other phrases. This notion has been
central to a number of approaches to grammar for some time, including
theories like dependency grammar (Hudson 1976, 1990) and HPSG
(Pollard and Sag 1987). More recently,
the statistical properties of associations between words, and
more particularly heads of phrases, has become an active area of
research (e.g. Chang, Luo, and Su 1992; Hindle and Rooth 1993).

The language model factors the statistical
derivation of a sentence with word string $W$ as follows:
\begin{dispquo}
$ P(W) = \sum_C P(C) \: P(W | C) $
\end{dispquo}
where $C$ ranges over relation graphs. The content model, $P(C)$, and
generation model, $P(W | C)$, are components of the overall
statistical model for spoken language translation given earlier.
This decomposition of $P(W)$ can be viewed as first deciding on
the content of a sentence, formulated as a set of relation edges
according to a statistical model for $P(C)$, and then deciding
on word order according to $P(W | C)$.

Of course, this decomposition simplifies the realities of language
production in that real
language is always generated in the context of some
situation $S$ (real or imaginary), so a more comprehensive
model would be concerned with $P(C | S)$, i.e. language
production in context. This is less important, however, in
the translation setting since we produce $C_t$ in the
context of a source relation graph $C_s$ and we
assume the availability of a model for $P(C_t | C_s)$.

\subsection{Content Derivation Model}

The model for deriving the relation graph of a phrase is taken
to consist of choosing a lexical head $h_0$ for the phrase
(what the phrase is `about') followed by a series of `node
expansion' steps.
An expansion step takes a node and chooses a possibly empty set
of edges (relation labels and ending nodes) starting from that node.
Here we consider only the case of relation graphs that are trees with
unordered siblings.

To start with, let us take the simplified case where a head word
$h$ has no optional or duplicated dependents (i.e. exactly one
for each relation).
There will be a set of edges
\begin{dispquo}
$ E(h) = \{r_1(h,w_1), \: r_2(h,w_2) \: \dots \: r_k(h,w_k)\} $
\end{dispquo}
corresponding to the local tree rooted at $h$ with
dependent nodes $w_1 \dots w_k$.
The set of relation edges for the entire derivation is the union
of these local edge sets.

To determine the probability of deriving a relation graph $C$
for a phrase
headed by $h_0$ we make use of parameters (`dependency parameters')
\begin{dispquo}
$ P(r(h,w) | h,r) $
\end{dispquo}
for the probability, given a node $h$ and a relation $r$, that $w$
is an $r$-dependent of $h$. Under the assumption
that the dependents of a head are chosen independently from each
other, the probability of deriving $C$ is:
\begin{dispquo}
$ P(C)= \: P(Top(h_0))
             \: \prod_{r(h,w) \in C}
                \: P(r(h,w) | h,r) $
\end{dispquo}
where $P(Top(h_0))$ is the probability of choosing $h_0$ to
start the derivation.

If we now remove the assumption made earlier that there is exactly
one $r$-dependent of a head, we need to elaborate the derivation
model to include choosing the number of such dependents.
We model this by parameters
\begin{dispquo}
$ P(N(r,n) | h) $
\end{dispquo}
that is, the probability that head $h$ has $n$ $r$-dependents.
We will refer to this probability as a `detail parameter'.
Our previous assumption
amounted to stating that this was always 1 for
$n=1$ or for $n=0$. Detail parameters allow
us to model, for example, the number of adjectival modifiers of
a noun or the `degree' to which a particular argument of a verb
is optional. The probability of an expansion of $h$ giving
rise to local edges $E(h)$ is now:
\begin{dispeqn}
\lefteqn{ P(E(h)|h) = }\\
& \prod_r \: P(N(r,n_r)|h)  \:
    \: k(n_r) \: \prod_{1 \leq i \leq n_r} \: P(r(h,w^r_i) | h,r).
\end{dispeqn}
where $r$ ranges over the set of relation labels and $h$ has
$n_r$ $r$-dependents $w^r_1 \dots w^r_n$. $k(n_r)$ is a
combinatoric constant for taking account of the fact that we are not
distinguishing permutations of the dependents (e.g. there are
$n_r!$ permutations of the  $r$-dependents of $h$ if these
dependents are all distinct).

So if $h_0$ is the root of a tree $C$, we have
\begin{dispquo}
$ P(C)= P(Top(h_0)) \: \prod_{h \in heads(C)} \: P(E_{C}(h)|h) $
\end{dispquo}
where $heads(C)$ is the set of nodes in $C$ and $E_{C}(h)$ is
the set of edges headed by $h$ in $C$.

The above formulation is only an
approximation for relation graphs that are not trees
because the independence assumptions
which allow the dependency parameters to be simply multiplied
together no longer hold for the general case.
Dependency graphs with cycles do arise as the
most natural analyses of certain linguistic constructions, but
calculating their probabilities on a node by node basis as above may
still provide probability estimates that are accurate enough for
practical purposes.

\subsection{Generation Model}

We now return to the generation model $P(W | C)$. As mentioned
earlier, since $C$ includes the words in $W$ and
a set of relations between them, the generation model is
concerned only with surface order. One possibility is to
use `bi-relation' parameters for the probability that
an $r_i$-dependent immediately follows an $r_j$-dependent.
This approach is problematic for our overall
statistical model because such parameters are not independent
from the `detail' parameters specifying the number of $r$-dependents
of a head.

We therefore adopt the use of `sequencing' parameters, these
being probabilities of particular orderings of dependents
given that the multiset of dependency relations is known.
We let the identity relation $e$ stand for the head itself.
Specifically, we have parameters
\begin{dispquo}
$ P(s|M(s)) $
\end{dispquo}
where $s$ is a sequence of relation labels including an
occurrence of $e$ and $M(s)$ is the multiset for this sequence.
For a head $h$ in a relation graph $C$, let $s_{WCh}$ be the sequence
of dependent relations induced by a particular word string $W$
generated from $C$. We now have
\begin{dispquo}
$ P(W|C) = \prod_{h\in W}(\prod_r\frac{1}{k(n_r)})
            P(s_{WCh}|M(s_{WCh})) $
\end{dispquo}
where $h$ ranges over all the heads in $C$, and $n_r$ is the
number of occurrences of $r$ in $s_{WCh}$,
assuming that all orderings of $n_r$-dependents are equally likely.
We can thus use these sequencing parameters directly in our overall
model.

To summarize, our monolingual models are specified by:
\begin{itemize}
\item topmost head parameters  $P(Top(h))$
\item dependency parameters $P(r(h,w)|h,r)$
\item detail parameters $P(N(r,n)|h)$
\item sequencing parameters $P(s|M(s))$
\end{itemize}

The overall model splits the contributions of content $P(C)$
and ordering $P(W|C)$. However, we may also want a model
for $P(W)$, for example for pruning speech recognition hypotheses.
Combining our content and ordering models we get:
\begin{dispeqn}
\lefteqn{ P(W) = \sum_C \: P(C) \: P(W | C)} \\
& = \sum_C P(Top(h_C)) \: & \prod_{h \in W} \: P(s_{WCh}|h) \\
&                         & \prod_{r(h,w) \in E_{C}(h)} \: P(r(h,w)|h,r)
\end{dispeqn}
The parameters $P(s|h)$ can be derived by combining sequencing
parameters with the detail parameters for $h$.

\section{Translation Model}
\label{tm}

\subsection{Mapping Relation Graphs}

As already mentioned, the translation model defines mappings
between relation graphs $C_s$ for the source language
and $C_t$ for the target language.
A direct (though incomplete) justification of translation
via relation graphs may be based on a
simple referential view of natural language semantics.
Thus nominals and their modifiers pick out entities
in a (real or imaginary) world, verbs and their modifiers
refer to actions or events in which the entities participate
in roles indicated by the edge relations. Under
this view, the purpose of the translation mapping is to
determine a target language relation graph that
provides the best approximation to the referential function
induced by the source relation graph. We call this
approximating referential equivalence.

This referential view of semantics is not adequate for
taking account of much of the complexity of natural language
including many aspects of quantification, distributivity
and modality.
This means it cannot capture some of the subtleties that
a theory based on logical equivalence might be expected
to. On the other hand, when we proposed a logic based
approach as our qualitative model, we had to restrict
it to a simple first order logic anyway for computational
reasons, and even then it did not appear to be practical.
Thus using the
more impoverished lexical relations representation may
not be costing us much in practice.

One aspect of the representation that is particularly useful
in the translation application is its convenience for partial
and/or incremental representation of content -- we can refine
the representation by the addition of further edges. A fully
specified denotation of the meaning of a sentence is rarely
required for translation, and as we pointed out when discussing
logic representations, a complete specification may not have
been intended by the speaker.
Although we have not provided
a denotational semantics for sets of relation edges,
we anticipate that this will be possible along the lines
developed in monotonic semantics (Alshawi and Crouch 1992).

\subsection{Translation Parameters}

To be practical, a model for $P(C_t|C_s)$ needs to decompose
the source and target graphs $C_s$ and $C_t$ into subgraphs
small enough that subgraph translation parameters can be estimated.
We do this with
the help of `node alignment relations' between the nodes of
these graphs. These alignment relations are similar in some
respects to the alignments used by Brown et al. (1990)
in their surface translation model. The translation probability
is then the sum of probabilities over different alignments $f$:
\begin{dispquo}
$ P(C_t|C_s) = \sum_{f}{P(C_t,f|C_s)} $.
\end{dispquo}
There are different ways to model $P(C_t,f|C_s)$ corresponding
to different kinds of alignment relations and different
independence assumptions about the translation mapping.

For our quantitative design, we adopt a simple model
in which lexical and relation (structural) probabilities are
assumed to be independent. In this model the alignment relations
are functions from the word occurrence nodes of $C_t$ to
the word occurrences of $C_s$. The idea is that $f(v_j)=w_i$ means
that the source word occurrence $w_i$ `gave rise' to the target word
occurrence $v_j$.
The inverse relation $f^{-1}$ need not be a function,
allowing different numbers of words in the source and target sentences.

We decompose $P(C_t,f|C_s)$ into `lexical' and `structural'
probabilities as follows:
\begin{dispquo}
$ P(C_t,f|C_s) = P(N_t,f|N_s)  P(E_t|N_t,f,C_s) $
\end{dispquo}
where $N_t$ and $N_s$ are the node sets for $C_t$ and $C_s$
respectively, and $E_t$ is the set of edges for the target graph.

The first factor $P(N_t,f|N_s)$ is the lexical component in that
it does not take into account any of the relations in the
source graph $C_s$. This lexical component is
the product of alignment probabilities for each node of $N_s$:
\begin{dispeqn}
\lefteqn{P(N_t,f|N_s)=}\\
& & \prod_{w_i \in N_s}
    P(f^{-1}(w_i)= \{v_i^1 \dots v_i^n\} | w_i).
\end{dispeqn}
That is, the probability that $f$ maps exactly the (possibly empty)
subset $\{v_i^1 \dots v_i^n\}$ of $N_t$ to $w_i$.
These sets are assumed to be disjoint for different source
graph nodes, so we can replace the factors in the above product
with parameters:
\begin{dispquo}
$ P(M | w) $
\end{dispquo}
where $w$ is a source language word and $M$ is a multiset of
target language words.

We will derive a target set of edges $E_t$ of $C_t$ by $k$ derivation
steps which partition the set of source edges $E_s$ into
subgraphs $S_1 \dots S_k$. These subgraphs give rise to
disjoint sets of relation edges $T_1 \dots T_k$ which together
form $E_t$.
The structural component of our translation model will
be the sum of derivation probabilities for such an edge set $E_t$.

For simplicity, we assume here that the source graph $C_s$ is
a tree. This is consistent with our earlier assumptions about the
source language model. We take our partitions of the source
graph to be the edge sets for local trees. This ensures that
the the partitioning is deterministic so the probability of
a derivation is the product of the probabilities of derivation steps.
More complex models with larger partitions rooted at a node are
possible but these require additional parameters for partitioning.
For the simple model it remains to specify derivation step
probabilities.

The probability of a derivation step is given by parameters of
the form:
\begin{dispquo}
$ P(T'_i|S'_i,f_i) $
\end{dispquo}
where $S'_i$ and $T'_i$ are unlabeled graphs and $f_i$ is a node
alignment function from $T'_i$ to $S'_i$. Unlabeled graphs are
just like our relation edge graphs except that the nodes are not
labeled with words (the edges still have relation labels).
To apply a derivation step we need a notion of graph matching
that respects edge labels:
$g$ is an isomorphism (modulo node labels) from a graph $G$ to
a graph $H$ if $g$ is a one-one and onto function from the
nodes of $G$ to the nodes of $H$ such that
\begin{dispquo}
$ r(a,b) \in G $ iff $ r(g(a),g(b)) \in H $.
\end{dispquo}

The derivation step with parameter $P(T'_i|S'_i,f_i)$ is
applicable to the source edges $S_i$, under the alignment $f$,
giving rise to the target edges $T_i$ if (i) there is an
isomorphism $h_i$ from $S'_i$ to $S_i$ (ii) there is
an isomorphism $g_i$ from $T_i$ to $T'_i$ (iii) for any node $v$
of $T_i$ it must be the case that
\begin{dispquo}
$ h_i(f_i(g_i(v))) = f(v) $.
\end{dispquo}
This last condition ensures that the target graph partitions
join up in a way that is compatible with the node alignment $f$.

The factoring of the translation model into these lexical and
structural components means that it will overgenerate because
these aspects are not independent in translation between
real natural languages. It is therefore appropriate to filter
translation hypotheses by rescoring according to
the version of the overall statistical model that included the
factors $P(C_t)P(C_s|C_t)$ so that the target
language model constrains the output of the translation model.
Of course, in this
case we need to model the translation relation in the `reverse'
direction. This can be done in a parallel
fashion to the forward direction described above.

\section{Conclusions}
\label{compare}

Our qualitative and quantitative models have a similar overall
structure and there are clear parallels between the factoring
of logical constraints and statistical parameters, for
example monolingual postulates and dependency parameters,
bilingual postulates and translation parameters.
The parallelism would have been closer if we had adopted ID/LP
style rules (Gazdar et al. 1985) in the qualitative model.
However, we argued in section~\ref{logical} that our qualitative
model suffered from lack of robustness, from having only the
crudest means for choosing between competing hypotheses, and
from being computationally intractable for large vocabularies.

The quantitative model is in a much better position to cope with
these problems. It is less brittle
because statistical associations have replaced constraints
(featural, selectional, etc.) that must be satisfied exactly.
The probabilistic models give us a systematic and well motivated
way of ranking alternative hypotheses. Computationally, the
quantitative model lets us escape from the
undecidability of logic-based reasoning.
Because this model is highly lexical, we
can hope that the input words will allow effective pruning by
limiting the number of search paths having significantly high
probabilities.

We retained some of the basic assumptions about the structure
of language when moving to the quantitative model. In particular,
we preserved the notion of hierarchical phrase structure.
Relations motivated by dependency grammar made it possible to
do this without giving up sensitivity to lexical collocations
which underpin simple statistical models like N-grams.
The quantitative model also reduced overall complexity in terms
of the sets of symbols used. In addition to words, it only
required symbols for dependency relations, whereas the qualitative
model required symbol sets for linguistic categories and features,
and a set of word sense symbols. Despite their apparent importance
to translation, the quantitative system can avoid the use of word
sense symbols (and the problems of granularity they give rise to)
by exploiting statistical associations between words in the target
language to filter implicit sense choices.

Finally, here is a summary of our reasons for combining statistical
methods with dependency representations in our language and
translation models:
\begin{itemize}
\item inherent lexical sensitivity of dependency representations,
      facilitating parameter estimation;
\item quantitative preference based on probabilistic derivation
      and translation;
\item incremental and/or partial specification of the content of
      utterances, particularly useful in translation;
\item decomposition of complex utterances through recursive
      linguistic structure.
\end{itemize}
These factors suggest that dependency grammar will play an
increasingly important role as language processing systems seek to
combine both structural and collocational information.

\section*{Acknowledgements}

I am grateful to Fernando Pereira, Mike Riley, and Ido Dagan
for valuable discussions on the issues addressed in this paper.
Fernando Pereira and Ido Dagan also provided helpful comments
on a draft of the paper.

\section*{References}

\begin{reverseindent}

\item\pagebreak[3]
Alshawi, H., D. Carter, B. Gamback and M. Rayner. 1992.
``Swedish-English QLF Translation''. In H. Alshawi (ed.)
{\it The Core Language Engine}, Cambridge, Mass.: MIT Press.

\item\pagebreak[3]
Alshawi, H. and R. Crouch. 1992. ``Monotonic Semantic Interpretation''.
{\it Proceedings of the 30th Annual Meeting of the Association for
Computational Linguistics}, Newark, Delaware.

\item\pagebreak[3]
Alshawi, H. and D. Carter. 1994. ``Training and Scaling Preference
Functions for Disambiguation''. To appear in {\it Computational
Linguistics}.

\item\pagebreak[3]
Brill, E. 1993. ``Automatic Grammar Induction and Parsing Free
Text: A Transformation-Based Approach''.{\it Proceedings of the 31st
Annual Meeting of the Association for Computational Linguistics},
259--265.

\item\pagebreak[3]
Brown, P., J. Cocke, S. Della Pietra, V. Della Pietra, F. Jelinek,
J. Lafferty, R. Mercer and P. Rossin. 1990. ``A Statistical Approach
to Machine Translation''. {\it Computational Linguistics} 16:79--85.

\item\pagebreak[3]
Chang, J., Y. Luo, and K. Su. 1992. ``GPSM: A Generalized Probabilistic
Semantic Model for Ambiguity Resolution''. {\it Proceedings of the 30th
Annual Meeting of the Association for Computational Linguistics},
177--192.

\item\pagebreak[3]
Chang, J., K. Su. 1993. ``A Corpus-Based Statistics-Oriented
Transfer and Generation Model for Machine Translation''.
{\it Proceedings of the 5th International Conference on Theoretical and
Methodological Issues in Machine Translation}.

\item\pagebreak[3]
Dagan I. and A. Itai. 1994. ``Word Sense Disambiguation Using a
Second Language Monolingual Corpus''. To appear in {\it Computational
Linguistics}.

\item\pagebreak[3]
Dagan, I., S. Marcus and S. Markovitch. 1993. ``Contextual Word
Similarity and Estimation from Sparse Data''.
{\it Proceedings of the 31st meeting of the
Association for Computational Linguistics}, ACL, 164--171.

\item\pagebreak[3]
Gazdar, G., E. Klein, G.K. Pullum, and I.A.Sag. 1985.
{\it Generalised Phrase Structure Grammar}. Oxford: Blackwell.

\item\pagebreak[3]
Hindle, D. and M. Rooth. 1993. ``Structural Ambiguity and Lexical
Relations''. {\it Computational Linguistics} 19:103--120.

\item\pagebreak[3]
Hobbs, J.R., M. Stickel, P. Martin and D. Edwards. 1988.
``Interpretation as Abduction'', Proceedings of the 26th
Annual Meeting of the Association for Computational Linguistics,
Buffalo, New York, 95-103.

\item\pagebreak[3]
Hudson, R.A. 1984. {\it Word Grammar}. Oxford: Blackwell.

\item\pagebreak[3]
Isabelle, P. and E. Macklovitch. 1986. ``Transfer and MT
Modularity'', {\it Eleventh International Conference on Computational
Linguistics}, Bonn, 115--117.

\item\pagebreak[3]
Jelinek, F., R.L. Mercer and S. Roukos. 1992. ``Principles of Lexical
Language Modeling for Speech Recognition''.
In S. Furui and M.M. Sondhi (eds.), {\it Advances in Speech Signal
Processing}, New York: Marcel Dekker Inc.

\item\pagebreak[3]
Mellish, C.S. 1988. ``Implementing Systemic Classification by
Unification''.  {\it Computational Linguistics} 14:40--51.

\item\pagebreak[3]
McCord, M. 1988. ``A Multi-Target Machine Translation System''.
Proceedings of the International Conference on Fifth Generation
Computer Systems, Tokyo, Japan, 1141--1149.

\item\pagebreak[3]
Pereira, F., N. Tishby and L. Lee. 1993. ``Distributional Clustering
of English Words''. {\it Proceedings of the 31st meeting of the
Association for Computational Linguistics}, ACL, 183--190.

\item\pagebreak[3]
Pollard, C.J. and I.A. Sag. 1987. {\it Information Based Syntax
and Semantics: Volume I --- Fundamentals}. CSLI Lecture Notes,
Number 13. Center for the Study of Language and Information,
Stanford, California.

\item\pagebreak[3]
Rayner, M. and H.  Alshawi.  1992.  ``Deriving Database Queries from
Logical Forms by Abductive Definition Expansion''. Proceedings
of the Third Conference on Applied Natural Language Processing,
Trento, Italy.

\item\pagebreak[3]
Richard, M.D. and R.P. Lippmann. 1991. ``Neural Network Classifiers
Estimate Bayesian {\it a posteriori} Probabilities''.
{\it Neural Computation} 3:461--483.

\item\pagebreak[3]
Shieber, S.M. 1986. {\it An Introduction to Unification-Based
Approaches to Grammar}.  CSLI Lecture Notes,
Number 4. Center for the Study of Language and
Information, Stanford, California.

\item\pagebreak[3]
Smajda, F. and K. McKeown. 1990. ``Automatically Extracting
and Representing Collocations for Language Generation''.
In {\it Proceedings of the 28th Annual Meeting  of the
Association for Computational Linguistics}, Pittsburgh.

\item\pagebreak[3]
Taylor, L., C. Grover, and E.J. Briscoe. 1989. ``The Syntactic
Regularity of English Noun Phrases''. Proceedings of the 4th
European ACL Conference, 256--263.

\item\pagebreak[3]
Weaver, W. 1955. ``Translation''. In W. Locke and A. Booth (eds.),
{\it Machine Translation of Languages},
Cambridge, Mass.: MIT Press.

\end{reverseindent}

\end{document}